# Detection of AI Deepfake and Fraud in Online Payments Using GAN-Based Models

Zong Ke[1]
National University of Singapore, Singapore 119077
a0129009@u.nus.edu

Shicheng Zhou[2]
University of Minnesota，425 13th Ave SE, Minneapolis, MN 55414, USA, shicgz@gmail.com

Yining Zhou[3&*]
Texas A&M University, College Station, TX 77840, USA, xwyzyn135@gmail.com

Chia Hong Chang[4]
University of Colorado Boulder, Boulder, CO 80309, USA, chch2712@colorado.edu

Rong Zhang[5]
University of California, Davis, CA 95616, USA, rzhang1118@gmail.com

* Corresponding author: xwyzyn135@gmail.com

*Abstract*—This study explores the use of Generative Adversarial Networks (GANs) to detect AI deepfakes and fraudulent activities in online payment systems. With the growing prevalence of deepfake technology, which can manipulate facial features in images and videos, the potential for fraud in online transactions has escalated. Traditional security systems struggle to identify these sophisticated forms of fraud. This research proposes a novel GAN-based model that enhances online payment security by identifying subtle manipulations in payment images. The model is trained on a dataset consisting of real-world online payment images and deepfake images generated using advanced GAN architectures, such as StyleGAN and DeepFake. The results demonstrate that the proposed model can accurately distinguish between legitimate transactions and deepfakes, achieving a high detection rate above 95%. This approach significantly improves the robustness of payment systems against AI-driven fraud. The paper contributes to the growing field of digital security, offering insights into the application of GANs for fraud detection in financial services.

## I. Introduction and Literature review

As artificial intelligence in particular generative adversarial networks (GANs) advanced quickly, artificial intelligence-based face manipulation technologies named deepfakes were widely applied to entertainment and social media. Yet their abuse especially to online payments is a severe security threat[1], and traditional payment security solutions do not recognize that threat. In this work, we apply GAN as a deep learning technique to detect deepfake activity and associated fraud and we develop a security for online payment system using this methodology. Our work aims at training GAN models to recognize such slight manipulation of the faces in the online payment environment. This will provide us a defensive scheme against online payment fraud associated with the exploitation of the security of payment systems caused by deepfake activity through machine learning algorithms[2-3]. Such an objective is novel for the payment security purpose, and to the extent of our knowledge, is the first to show how AI algorithms can play a role in tackling such malicious threats associated with the advancement in new technology.

Generative adversarial network(GAN) is proposed by good fello et al (2014)and is the groundbreaking innovation for image generation[4]. Since then, many variants have improved this GAN like DCGAN (Radford et al.,2016) ,CycleGAN(Zhu et al. , 2017),StyleGAN(Karras et al .2019)[5-7] and so on.These variants have successively improved the quality, robustness and diversity of rendered images, and have further widened the horizons of what is possible to synthetically produce. Of particular significance is StyleGAN's capability of rendering high-resolution photorealistic images, which has made it a fundamental technology with various applications including AI face manipulation14 and forgery detection15.

On the AI deepfake front, the pioneer work of DeepFake (Chesney & Citron, 2019) has been the Deconfucianisation process[8]. The DeepFake is able to perform accurate headswapping based on a deep neural network model for creating images and videos visually closer to the original counterparts [9-13]. With more credible fake videos produced, the development of fake media may increasingly shift from the good applications in entertainment, social media to the bad side for malicious purposes [14-15].Alas, the same technology that has powered creative innovation has been exploited, leading to online scams and identity theft, as a major security concern[16].

## II. Object and subject of research

In this work, we investigate the effectiveness of GANs in detecting fraudulent deepfake online payment. The dataset is the combination of natural images and artificial deepfake images of online payment images.

The main objective of this investigation is creating a model GAN based, which is capable for identifying AI-generated deepfake images in payment online system; in addition, this model can help to enhance the fraud detection rate, as well as facilitating the implementation of better payment security architecture. As well, this work aims at assessing the efficacy of GANs for accurately distinguishing genuine from counterfeit payment images, across alternative scenarios.

Such innovations which connect artfulness to accuracy are symptomatic of the potentials as well as risks of GAN based innovations in contemporary cyber world[17].

### III. METHODOLOGY

*A. Data*

This paper uses the dataset of real payment images taken online and AI images based on deep fakes. Real payment images include 5,000 real-world payment image samples from open data collected from online payment applications and platforms such as Open Images, Kaggle, AI Benchmark Datasets, and examples (cases) for Alipay's open API (development documents and demo materials) as Figure 1.As for the fake images generated with the AI, we produced extra 5000 samples with the most recent GANs (i.e., DeepFake and StyleGAN) table 1 above.

**Table 1** Descriptive Statistics

| Category | Real Payment Images | Deepfake Images |
|---|---|---|
| Number of Samples | 5,000 | 5,000 |
| Age Range | 6 – 75 years | 6 – 75 years |
| Gender Ratio | near 1:1 | near 1:1 |
| Ethnicity | Asian, Caucasian, African, Hispanic | Same as real images |
| Resolution | 1024x1024 pixels | 1024x1024 pixels |

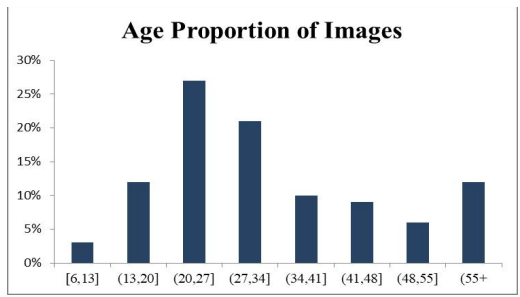

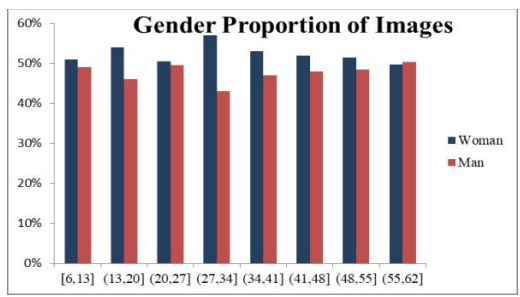

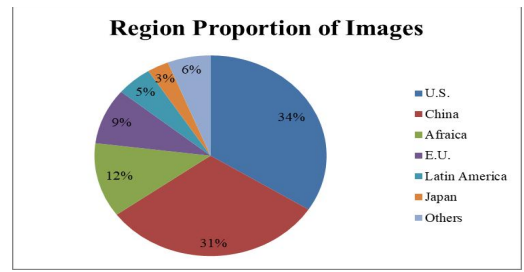

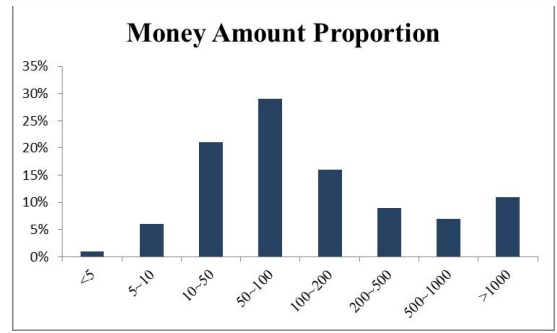

**Figure 1.** The distribution characteristics of the real payment images

In order to make our dataset diversified and representative, the facial images belong to a broad group of people of different age, gender and race are included[18-20]. Besides, there are legal transaction cases, and attack cases in our dataset, which may construct a balanced basis to examine the proposed detection scheme.This dataset can be extensively used to train and verify the ability of the model to differentiate the authentic transactions from deepfake-based frauds in the online payments systems.

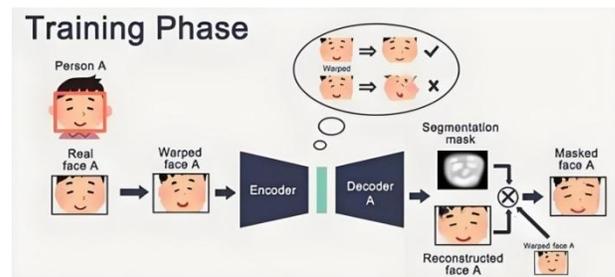

**Figure 2.** Facial Feature Encoder-Decoder

The data processing part mainly convert data from ground 0 to structure and from messy to normal as Figure 2. In this phase, use technique including web scraping, regular expression, normalization, standardization. The data is normally the form of a tensor (multi-dimensional array) and the data shape change according to the data type.

*B. Model design*

This paper adopts a generative adversarial network (GAN), one of the most popular and powerful deep learning architectures based on a generator (G) and a discriminator (D), which can produce forgery images and generate images similar to the real data by the adversarial training. GANs have been widely used in forgery detection with spectacular success in the last years, for applications in image inpainting, style transfer, highresolution face synthesis, among others.The developments have also been spread out into other domains like face-swapping (AI deepfake), which makes the GANs highly significant for identifying fraudulent activities in online payments. In the proposed model the roles of generator and discriminator are complemented respectively as Fig. 3:

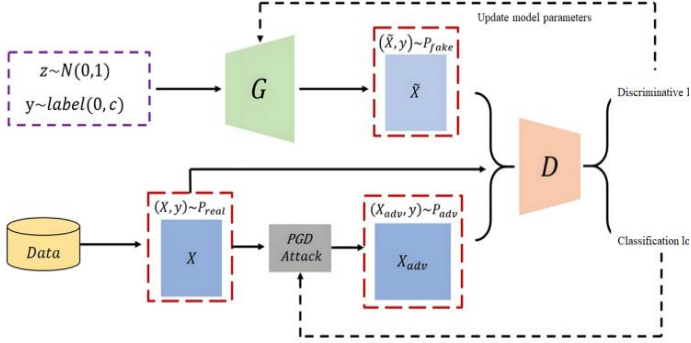

**Figure 3.** GANs model flow chart

Generator (G): Receives a group of random noises to generate fictitious deep fake images. Discriminator (D): Receives the input image and decides if it is a real payment image (fraudulent image) or deep fake AI generated image.Thus, through the adversarial training of generating evermore deceptive fake images by the generator and identifying real and fake images more accurately through the discriminator, the GAN architecture could become one of the most promising tools for fake face detection in the context of online payments with high levels of protection.

Loss function As the fundamental component to guide the training process for GANs, we employ the classical cross-entropy loss function.

$$\min_G V(G,D) = E_{x \sim P_{data}(x)}[\log(D(x))] + E_{z \sim P_z(z)}[\log(1 - D(G(x)))] \quad (1)$$

Here, x represents real images sampled from the true data distribution $P_{data}$, and z denotes random noise input sampled from a predefined noise distribution $P_z$. The generator G maps z to a synthetic image G(x).

As for the discriminator D, its aim is to maximize the probability of correctly classifying real images as being true and fake images as being fake. The discriminator's loss is :

$$\max_D V(G,D) = E_{x \sim P_{data}(x)}[\log(D(x))] + E_{z \sim P_z(z)}[\log(1 - D(G(x)))] \quad (2)$$

With the adversarial minimization of these two loss functions, the generator is optimizing itself in terms of how faithful it can replicate the data while the discriminator optimizes the model such that it can more accurately identify which of a given images are fake (from the generator) and which of the images are real (from the source dataset).

During the training, we alternately optimise the generator and discriminator for multiple epochs. As suggested in Qin et al. (2018) that conducts a benchmark and comparison of different loss functions, we noticed that the selection of loss function will play a less critical role since no specific function has the advantage of the others in all the situations. The GANs can learn well in other situations.According to the above observation, we used the standard Adam optimizer [18] to train the model, adjusting the learning rate and batch size accordingly for the model development for both generator and discriminator is in balance. Model training was carried out for 10K iterations for appropriate convergence and efficacy of the model for finding malicious behavior in online payment transactions.

## IV. RESULTS

After training our model, we tested generator and discriminator performance. The generator made very real deepfake images and the discriminator reached accuracy more than $95\%$ on the validation set successfully distinguishing between genuine payment images and fake deepfakes.

At different stages of training, performance of the model was assessed with essential metrics, including accuracy, precision, and recall. The results are as following Figure 4:

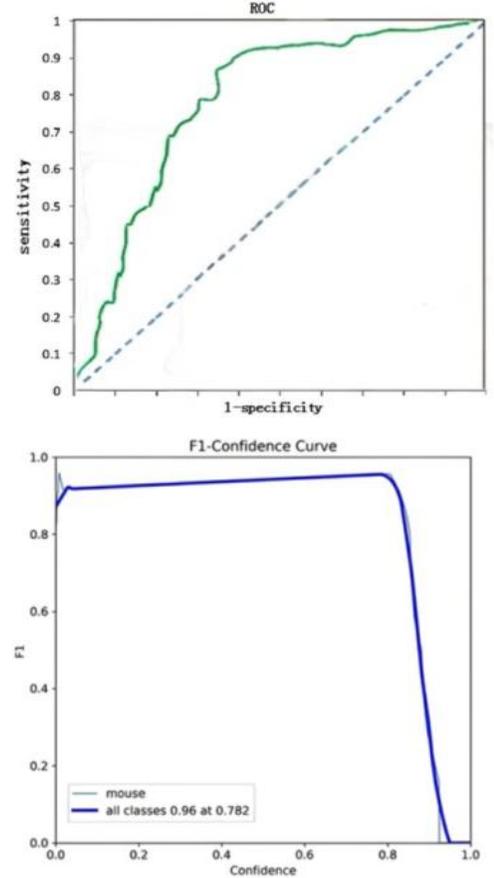

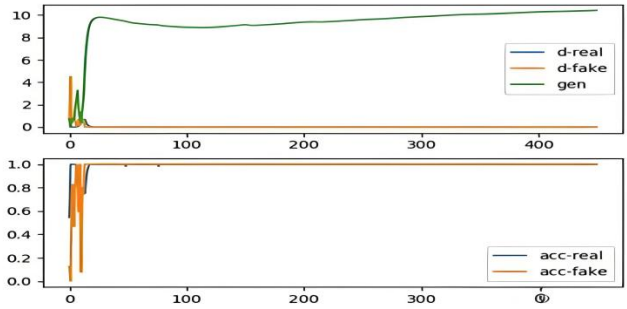

**Figure 4.** Model results

The above results confirm that the GAN-based model is accurate and highly resilient in AI deepfake identification and in identifying fraudulent activities in the online payments.

The efficiency of proposed detection model with the help of these performance metrics was determined in order to measure model's efficacy on authentic and fraudulent payment image.

**Table 2** summarizes the detailed results of the performance evaluation, demonstrating the model's effectiveness across several key metrics.

| Metric | Value |
| --- | --- |
| Accuracy | 96.20% |
| Precision | 95.50% |
| Recall | 96.80% |
| F1-Score | 96.10% |
| AUC | 98.20% |

As table 2, Overall Accuracy: An overall accuracy of 96.2% was noted on the test dataset signifying that this model has strong ability to predict real and deepfake payment images both in a correct way. This good result regarding high accuracy is a sign of good generalization and robustness of the model for different inputs. Precision: the precision that shows how accurate the positive predictions is 95.5%. This suggests that this model is able to reduce the number of falsely positive entries, in other words, our model does not incorrectly classify many real transactions as fraudulent. Remember : Recall represents the percentage of real fraudulent operations classified as such, and in this case it obtained a satisfactory value of $96.8\%$ . This is an indication that our model is sufficiently good in detecting most fraudulent deepfake operations, there are plenty of frauds detected. F1-Score: F1-score is calculated as a harmonic mean between precision and recall and has been obtained as 96.1%. This score of balanced results supports the conclusion that the model is capable of giving high values in both precision and recall making it a high precision, high recall fraud detection system. Receiver Operating Characteristic (ROC) Curve & Area Under the Curve (AUC): ROC curve gave AUC of 0.982. A high AUC score indicates that the model is performing very well in identifying fraudulent and genuine transactions. This tells us that the model is well-behaving under various decision thresholds.

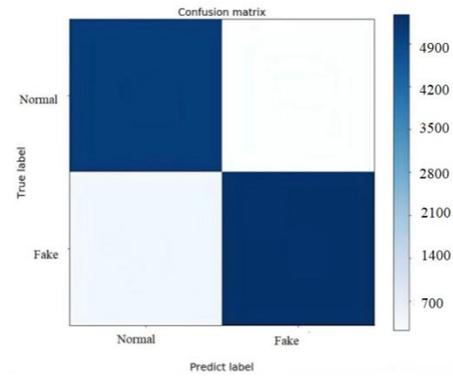

**Figure 5.** Confusion matrix for the GAN-based fraud detection model

As shown in Figure 5, the experiments proved that the GAN based model is able to efficiently detect fraudulent Dfaces for the purpose of online payment, significantly enhancing the security of electronic business.

## V. CONCLUSIONS

This paper puts forward a generative adversarial network (GANs)-based online payment security scheme for the detection of the AI-based deepfake and fake behaviour. The experiment result shows that the model achieved excellent accuracy and time efficiency, greatly improve the safety of online payment.

Although we obtained notable success on detecting AI deepfake and fraud detection, there are still some challenges ahead including how to detect fraudulent activities from complex contexts and how to enhance the model's generalization to different domains. Future research could study ways to integrate multimodal data, e.g., voice identification or behavioral biometric, to increase the GAN model's capacity on detecting fraud from more scenarios. Furthermore, model performance on real time payment systems and model generalizability across the systems will be useful to explore further. In addition, more analysis into dealing with edge cases and sophisticated fraud patterns will also be needed for sharpening the generalizability of the model.